\documentclass[letterpaper, 10 pt, journal]{IEEEtran}

\IEEEoverridecommandlockouts
\usepackage{comment}
\usepackage{amsmath} 
\usepackage{amssymb}  
\usepackage{graphicx}
\usepackage{amsfonts}
\usepackage{gensymb}
\usepackage{algorithmic}
\usepackage{textcomp}
\usepackage{xcolor}
\usepackage{graphicx}
\usepackage{dblfloatfix}    
\usepackage{pdfpages}

\usepackage{cite}
\makeatletter
\let\NAT@parse\undefined
\makeatother
\usepackage{hyperref}
\def\BibTeX{{\rm B\kern-.05em{\sc i\kern-.025em b}\kern-.08em
    T\kern-.1667em\lower.7ex\hbox{E}\kern-.125emX}}
    
\begin{document}

\title{Dynamic Locomotion Teleoperation of a Wheeled Humanoid Robot Reduced Model with a Whole-Body Human-Machine Interface}
\author{{Sunyu Wang$^1$ and Joao Ramos$^2$}
\thanks{$^1$The author was with the Department of Mechanical Science and Engineering at the University of Illinois at Urbana-Champaign, USA. The author is currently with the Robotics Institute at Carnegie Mellon University, USA. Corresponding author's contact: sunyuw@andrew.cmu.edu}
\thanks{$^2$The author is with the Department of Mechanical Science and Engineering at the University of Illinois at Urbana-Champaign, USA.}
\thanks{This work is supported by the National Science Foundation via grant IIS-2024775.}}

\maketitle

\begin{abstract} \\
Bilateral teleoperation provides humanoid robots with human planning intelligence while enabling the human to feel what the robot feels. It has the potential to transform physically capable humanoid robots into dynamically intelligent ones. However, dynamic bilateral locomotion teleoperation remains as a challenge due to the complex dynamics it involves. This work presents our initial step to tackle this challenge via the concept of wheeled humanoid robot locomotion teleoperation by body tilt. Specifically, we developed a force-feedback-capable whole-body human-machine interface (HMI), and designed a force feedback mapping and two teleoperation mappings that map the human's body tilt to the robot's velocity or acceleration. We compared the two mappings and studied the force feedback's effect via an experiment, where seven human subjects teleoperated a simulated robot with the HMI to perform dynamic target tracking tasks. The experimental results suggest that all subjects accomplished the tasks with both mappings after practice, and the force feedback improved their performances. However, the subjects exhibited two distinct teleoperation styles, which benefited from the force feedback differently. Moreover, the force feedback affected the subjects' preferences on the teleoperation mappings, though most subjects performed better with the velocity mapping. 
\end{abstract}


\section{Introduction}
\IEEEPARstart{S}{tate}-of-the-art autonomous humanoid robots have demonstrated remarkable dynamic motion capabilities in controlled environments \cite{Atlas}. However, when operating in unknown environments---such as navigating in disaster scenes and manipulating everyday objects in homes---autonomous humanoid robots tend to fall short of societies' expectations. 

This limitation exists largely because these robots' artificial brains lack the intelligence and intuition required for planning complex actions under uncertainties. On the other hand, humans possess such planning skills through motor learning. Hence, teleoperation, which supplies human movement information to robots as motion plans, appears promising in transforming physically capable humanoid robots into dynamically intelligent---and useful---ones. 

Humanoid robot teleoperation includes two sub-domains: 1) Manipulation teleoperation. 2) Locomotion teleoperation. Manipulation teleoperation typically involves upper limbs, and many related works concern fixed-base teleoperation systems and their kinematics alone \cite{Hauser_trina, Ciocarlie_teleoperation, my_ICRA_2021}. Locomotion teleoperation, however, typically involves lower limbs and needs to account for the system's dynamics due to the nature of human locomotion. This work aims to contribute to the specific area of dynamic humanoid robot locomotion teleoperation. 

\begin{figure}[t]
\centering
    \includegraphics[width = 0.999\linewidth]{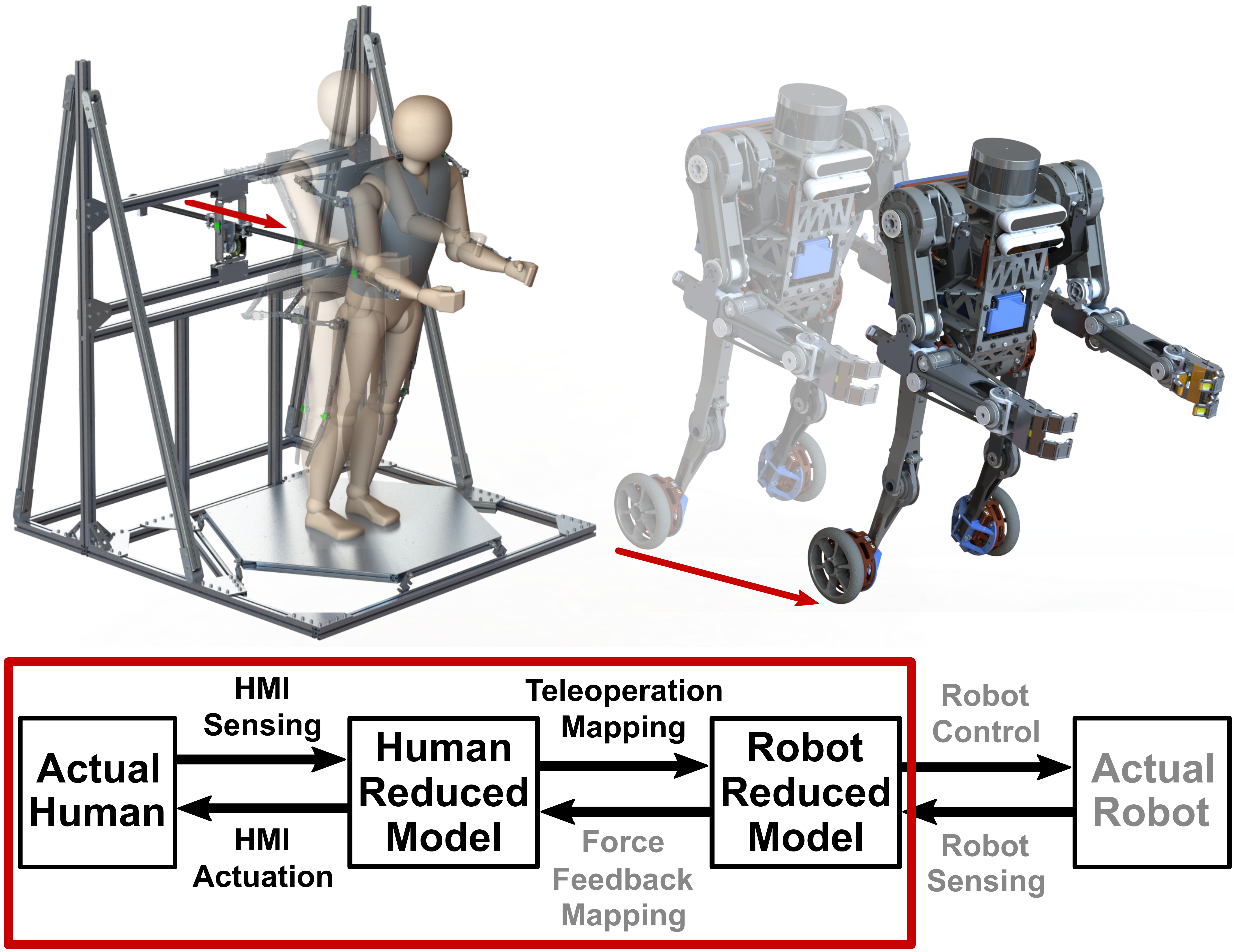}
    \caption{(Top) Design rendering of our envisioned teleoperation system, where a human uses a whole-body HMI to teleoperate a bi-wheeled humanoid robot's locomotion via body tilt. Images are not in scale. (Bottom) The four-stage bilateral teleoperation architecture. The red box indicates the scope of this work. Grey font color indicates the components not implemented in this work. }
    \label{fig:envisioned_system}
\end{figure}

Several existing works on humanoid robot locomotion teleoperation utilize a human-machine interface (HMI), a reduced model, and a teleoperation mapping \cite{iCub_2018, iCub_2019, MECHA_Unilateral, MECHA_Bilateral, Ramos_TRO_2018, Ramos_RAL_2018}. The HMI captures the human's information to generate the human reduced model. The teleoperation mapping then translates variables of the human reduced model into control commands for the robot reduced model. In this way, the teleoperation problem becomes a tractable four-stage tracking control problem, as in Fig. \ref{fig:envisioned_system}. 

In practice, commercial tetherless motion capture systems and omnidirectional treadmills have been popular HMIs because they provide useful human locomotion information and are readily available. The planar linear inverted pendulum (LIP) model has been a popular reduced model because it is one of the simplest models that encode the core dynamics of humanoid robot locomotion \cite{hof_LIP}. With these popular HMIs and LIP as the reduced model, teleoperation mappings designed to match the human's and the robot's zero-moment points (ZMPs) and divergent components of motion (DCMs) have produced meaningful locomotion behaviors of humanoid robots \cite{MECHA_Unilateral, iCub_2018, iCub_2019, multimode}. 

However, with such teleoperation methods, the human pilot cannot feel what the robot feels or comfortably control the robot's physical interaction with the environment. This is because the teleoperation is unilateral---the human sends information to the robot but receives little information from the robot. Unlike when a driver drives and feels a car inside it, the human pilot is physically away from the robot and cannot feel the robot's response to the human's command or the environment. In this regard, bilateral teleoperation with force feedback becomes valuable: it aims to allow the human pilot to feel what the robot feels in addition to synchronizing movements of the two. Due to the lack of available commercial hardware, most existing works on bilateral locomotion teleoperation of humanoid robots utilize custom-built HMIs. With LIP as the reduced model, ZMP- and DCM-based teleoperation and force feedback mappings, these works have achieved initial successes \cite{MECHA_Bilateral, haptic_walking, Ramos_Science_Robotics_2019, Ramos_TRO_2018}. 

Yet, the locomotion behaviors of those bilaterally teleoperated humanoid robots are preliminary. Some of them move quasi-statically, whereas others that are capable of dynamic motions cannot travel reliably for long distances. These limitations exist because walking is these robots' form of locomotion. The sophisticated dynamics of walking plus the human-robot coupled dynamics make dynamic bilateral teleoperation of walking and locomotion challenging. 

To tackle this challenge, we envision a bi-wheeled humanoid robot design for teleoperation, as shown in Fig. \ref{fig:envisioned_system}. The wheels simplify the locomotion teleoperation problem, and the humanoid design preserves the robot's anthropomorphism and intuitiveness of teleoperation. We developed a whole-body HMI with high backdrivability, bandwidth, and force capacity for bilateral teleoperation of our envisioned robot. The contribution of this work is twofold: 1) To introduce the HMI's design and performance. 2) To demonstrate and evaluate our concept of wheeled humanoid robot locomotion teleoperation by body tilt using the HMI via an experiment. 

Specifically, we employed LIP as the reduced model, and developed two teleoperation mappings that map the human's body tilt to the robot's velocity or acceleration. To prevent the human from falling during the teleoperation, we designed the HMI feedback force as a spring force proportional to the human's body tilt. Then, we conducted an experiment, where seven human subjects teleoperated a simulated LIP with the HMI to perform dynamic target tracking tasks. The experimental results suggest that all subjects accomplished the tasks after practice, and the force feedback generally improved their performances. However, the subjects exhibited two distinct teleoperation styles, which benefited from the force feedback differently. Moreover, the force feedback affected the subjects' preferences on the teleoperation mappings, though most subjects performed better with the velocity mapping. 

\begin{figure*}[t]
\centering
    \includegraphics[width = 0.999\linewidth]{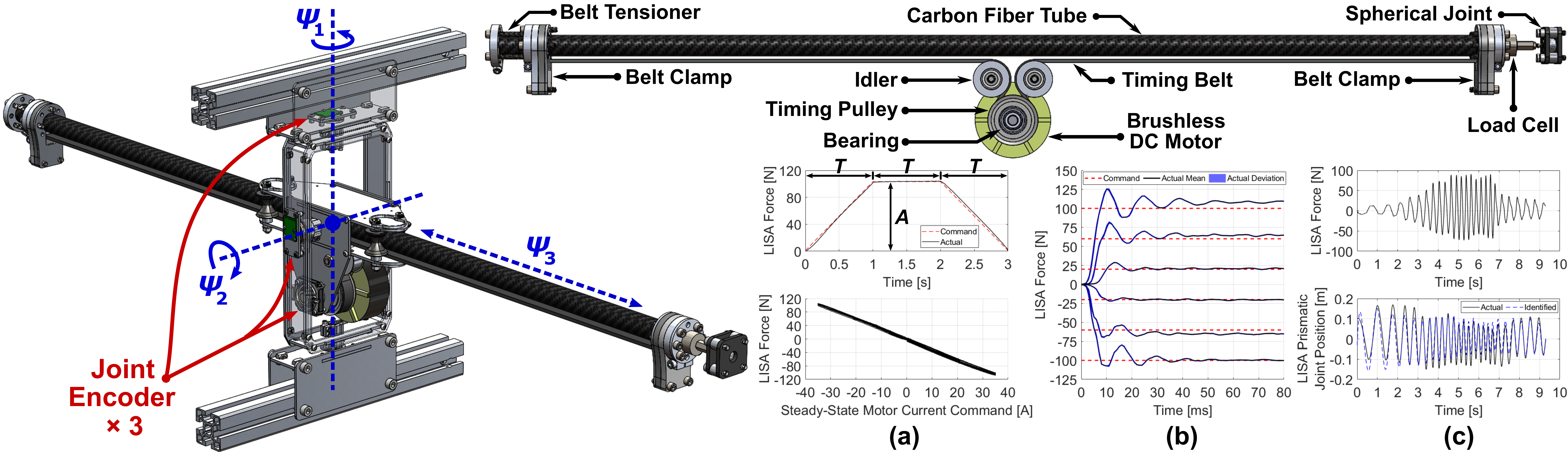}
    \caption{LISA's overall design, transmission topology, and performance. Images are not in scale. (a) (Top) Trapezoidal force command VS time profile in trapezoid test. Time interval $T \in \left\{ 1, 2, 3 \right\}$ s. Force amplitude $A \in \left\{ \pm 20, \pm 60, \pm 100 \right\}$ N, where tension is positive and compression negative. We performed five trials for each $T$-$A$ combination. (Bottom) Identified LISA force VS steady-state current profile. (b) Step responses. Step amplitudes are the same as the force amplitudes in the trapezoid test. We performed five trials for each step amplitude. (c) Chirp test force and joint position results. The identified joint position in the blue dashed line is the response obtained by entering the force VS time data to the identified transfer function model as input. }
    \label{fig:LISA}
\end{figure*}

\section{HMI Design and Performance}
The HMI we developed consists of three subsystems: 1) The linear sensor and actuator (LISA). 2) The forceplate. 3) The motion capture linkage. We introduced the motion capture linkage in our previous work on arm teleoperation \cite{my_ICRA_2021}. Hence, the following subsections will detail LISA's and the forceplate's designs and performances. 

\subsection{Linear Sensor and Actuator (LISA)}
LISA is the subsystem that senses the human pilot's center of mass (CoM) position and exerts feedback forces to the human. As shown in Fig. \ref{fig:LISA}, it is a three-degree-of-freedom (3-DoF) serial mechanism with two passive revolute joints on a gimbal and one actuated and backdrivable prismatic joint that contains the end-effector. A timing belt-pulley transmission converts the rotation of a Turnigy 9235-100 KV brushless DC motor to the end-effector's translation. The theoretical transmission ratio is $\frac{1000}{26}$ $\text{m}^{-1}$. The motor is current controlled by an ODrive V3.6 motor driver with a 500-Hz PWM command and a 40 V, 4 Ah lithium-ion battery. Three joint encoders sense the motor's and the gimbal's rotational axes at 1 kHz. A uniaxial tension/compression load cell senses the end-effector's actuation force at 7.58 kHz. Upon use, the gimbal is mounted to a fixed frame, and the end-effector is mounted to the human pilot's back at CoM height via a spherical joint and a vest. This mounting design allows unconstrained movement of the human's torso with a maximum swivel of about $90 \degree$. 

Empirically, a force of 50--60 N exerted at a human pilot's CoM would sufficiently perturb the human for humanoid robot teleoperation \cite{Ramos_TRO_2018, Ramos_RAL_2018}. To ensure an ample force capacity, we designed LISA's maximum actuation force to be 100 N. To evaluate LISA's actual force capacity and identify its actuation force VS motor current profile, we conducted a trapezoid test. First, we mounted LISA's gimbal ground and end-effector to the same fixed frame such that LISA's end-effector was perpendicular to the frame, and would apply a pure normal force to the frame when the prismatic joint was actuated. Second, we fed trapezoidal current command VS time profiles to LISA, and recorded the end-effector load cell reading and current command at 200 Hz. The result is a relatively linear curve with a maximum force of about 100 N at 35 A for both tension and compression, as shown in Fig. \ref{fig:LISA}.(a). 


After the trapezoid test, we programmed an open-loop force controller for LISA using the identified force-current plot's linear best-fit curve. Then, we changed the data logging frequency to 1 kHz, and conducted a step response test. Fig. \ref{fig:LISA}.(b) shows the results. The responses are relatively consistent, and the 100\% rise time is within 10 ms. The steady-state tracking error is acceptable because LISA's actuation force will be the feedback force for the human. A few Newtons of error should be negligible in this case. The relatively long settling time could be caused by LISA's mechanical compliance. 

Lastly, we conducted a chirp test to estimate the reflected inertia on LISA's prismatic joint. During the test, a human grasped LISA's end-effector, and back drove the prismatic joint in a sine-sweep manner for ten seconds, as shown in Fig. \ref{fig:LISA}.(c). The data were recorded at 1 kHz. We then loaded the data to the MATLAB System Identification Toolbox with a $2^{\text{nd}}$-order transfer function model. The result was approximately 1.3 kg reflected inertia, which is about $1\%$--$2\%$ of human body mass. Hence, we assume LISA's joint dynamics to be negligible compared with the human's dynamics. Furthermore, we measured the mass of all moving components on LISA's prismatic joint using a weight scale. The result was 0.655 kg. 

\subsection{Forceplate}
\begin{figure}[t]
\centering
    \includegraphics[width = 0.999\linewidth]{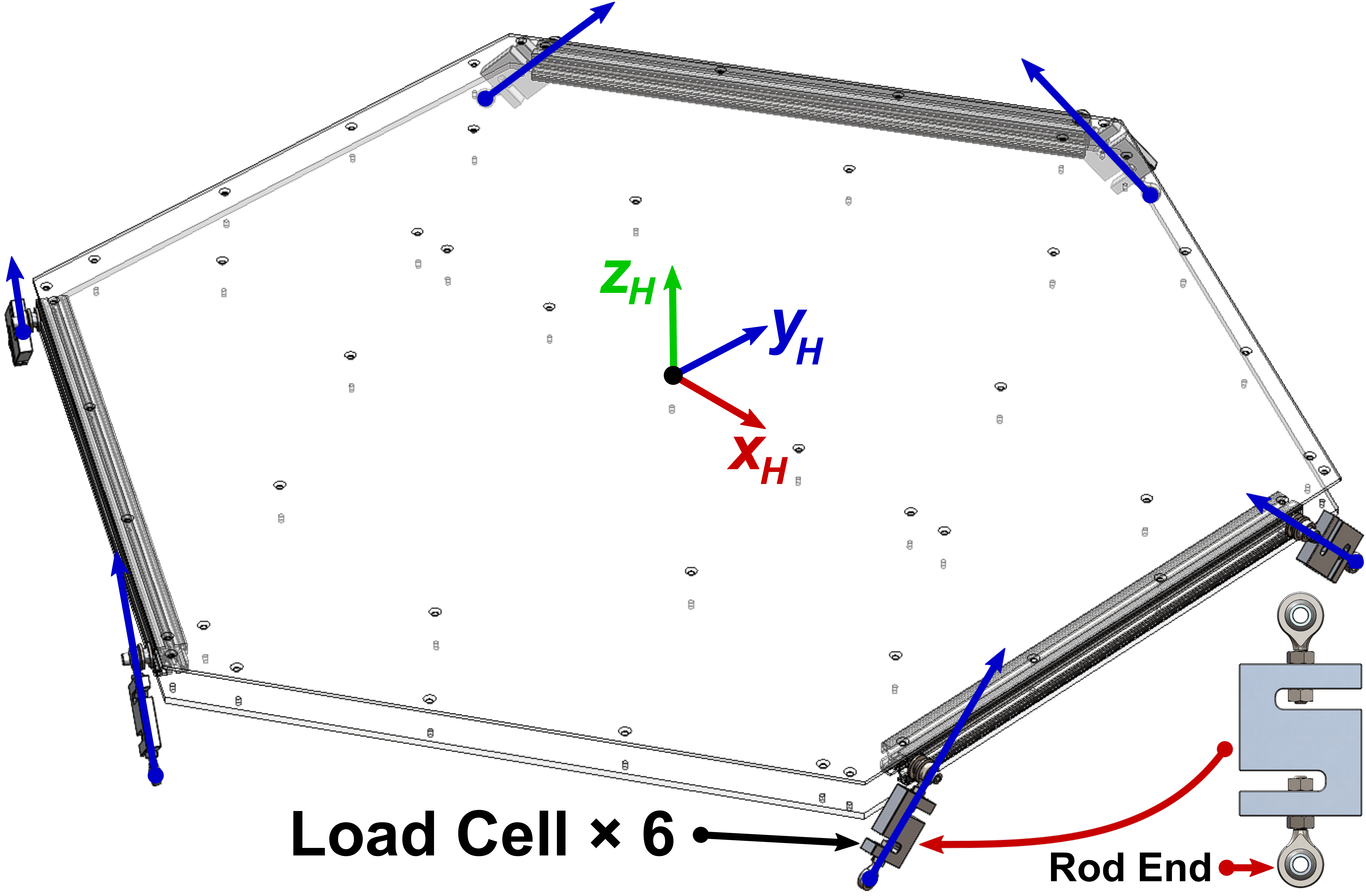}
    \caption{Forceplate's overall design, leg design, and coordinate frame definition. The forceplate's coordinate frame is the entire HMI's coordinate frame.}
    \label{fig:forceplate}
\end{figure}

The forceplate is the subsystem that senses the human pilot's ground reaction wrench and center of pressure (CoP) position. As shown in Fig. \ref{fig:envisioned_system} and Fig. \ref{fig:forceplate}, it consists of a rigid hexagonal platform connected to a fixed base frame via six legs. Each leg contains two spherical joints and a uniaxial tension/compression load cell operating at 7.58 kHz, so the subsystem is 6-DoF. However, these six DoFs are the six legs' axial rotations, which do not affect any leg's length. Hence, the platform has zero DoF and remains static upon external forces. Meanwhile, the forceplate's coordinate frame is the entire HMI's coordinate frame. Its origin is at the geometric center of the hexagonal platform's top surface. 

The forceplate's leg arrangement is identical to that of a Stewart platform, and the forceplate functions as a Stewart platform sensor \cite{stewart_sensor_1, stewart_sensor_2}. Specifically, since the hexagonal platform is static, the forceplate has a constant square Jacobian, which establishes a linear relationship between the six load cell readings and the wrench experienced by the platform:
\begin{align}
    \mathcal{F} = J^{-\intercal} f, 
\end{align}
where $\mathcal{F} \in \mathbb{R}^6$ is the wrench the platform experiences, $f \in \mathbb{R}^6$ is the column vector of the six load cell readings, and $J \in \mathbb{R}^{6 \times 6}$ is the Jacobian. Since $J$ is constant and invertible, $J^{-\intercal}$ can be directly obtained via a calibration process: applying six or more different known wrenches to the platform that excite all three axes of the forceplate's coordinate frame. Once the corresponding load cell readings are collected, a least-square matrix inversion would yield $J^{-\intercal}$. Lastly, the human's CoP position can be computed from the wrench itself. 


\section{LIP-Based Locomotion Teleoperation and Force Feedback Mappings}
The planar LIP model is a 1-DoF model where the pendulum's CoP controls its CoM, which translates at a constant height \cite{hof_LIP}. LIP's dynamical equation is: 
\begin{align} \label{LIP_dynamics}
    \ddot{x} \left( t \right) = \omega^2 \left( x \left( t \right) -  p \left( t \right) \right),
\end{align}
where $x(t)$ and $p(t)$ are the pendulum's CoM and CoP positions, respectively, and $\omega = \sqrt{g/h}$ is the natural frequency. $g$ and $h$ are gravitational acceleration and pendulum height, respectively, which are constant. Note that $x(t)$ is a continuous state while $p(t)$ is the input, which could be discontinuous. 

For locomotion teleoperation of our envisioned bi-wheeled humanoid robot, we assume that the planar LIP model represents the human and the robot in their respective sagittal planes. Physically, the LIP's CoM corresponds to the robot's CoM, and the LIP's CoP corresponds to the robot's wheel-ground contact point. For the human, the LIP's CoM and CoP are sensed by LISA and the forceplate, respectively. We designed two teleoperation mappings---feedback (FB) and feedforward (FF) mappings---that map the human LIP's tilt to the robot LIP's velocity or acceleration. With LISA's actuation capability, we also designed a force feedback mapping. The following subsections detail these mapping designs. 

\begin{figure}[t]
\centering
    \includegraphics[width = 0.999\linewidth]{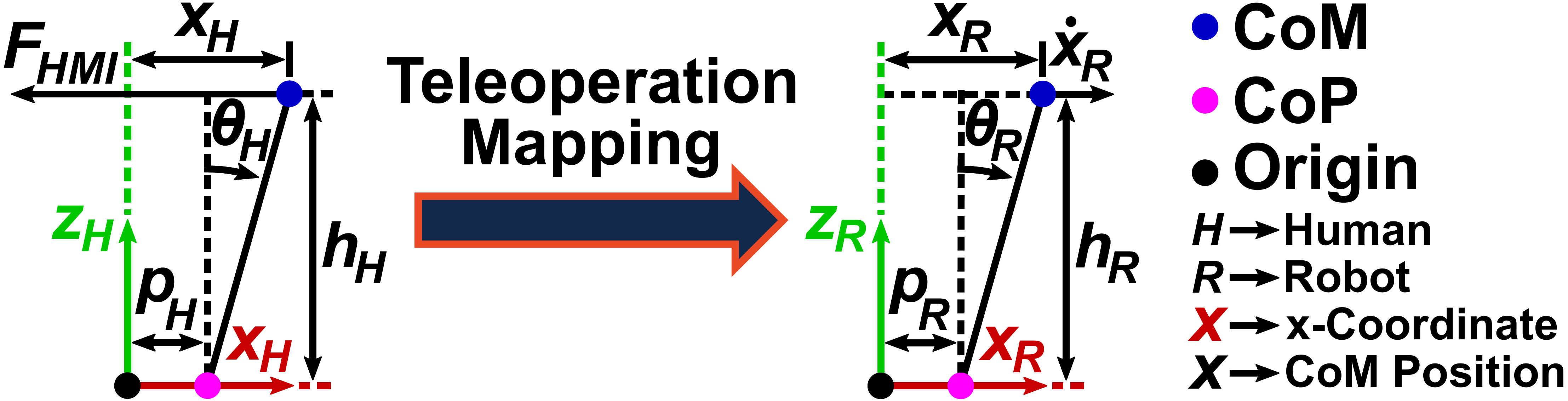}
    \caption{Schematics of the human LIP (left) and the robot LIP (right).}
    \label{fig:LIP}
\end{figure}

\subsection{Feedback Teleoperation Mapping}
FB mapping maps the human LIP's tilt to the robot LIP's CoM velocity. Intuitively, the human body acts as a control lever that commands the speed at which the robot should be travelling. Technically, the robot has a stabilizing feedback controller that accepts a velocity command. FB mapping translates the human's body tilt into that velocity command with a gain. The name ``feedback" comes from the design that the robot has its own feedback controller. 

Mathematically, the robot's feedback controller is: 
\begin{align} \label{p_R_FB}
    p_{R} = - \frac{\ddot{x}_{Rcmd}}{\omega_R^2} - \frac{2\zeta_R}{\omega_R} \left( \dot{x}_{Rcmd} - \dot{x}_R \right) - x_{Rcmd} + 2 x_R, 
\end{align}
where $p_R$ and $x_R$ are the robot LIP's CoP and CoM positions, respectively, $\omega_R$ is the robot LIP's natural frequency, and $\zeta_R$ is the controller's damping ratio. Subscript $``cmd"$ stands for ``command". Substituting (\ref{p_R_FB}) into the robot LIP's dynamical equation yields: 
\begin{equation}
    \left( \ddot{x}_{Rcmd} - \ddot{x}_{R} \right) + 2\zeta_R \omega_R \left( \dot{x}_{Rcmd} - \dot{x}_{R} \right) + \omega_R^2 \left( x_{Rcmd} - x_R \right) = 0, 
\end{equation}
which is the standard $2^{\text{nd}}$-order error dynamics. FB mapping, then, has the expression: 
\begin{align}
    \dot{x}_{Rcmd} = K_{FB} \left( x_H - p_H \right),
\end{align}
where $K_{FB}$ is the user-selected mapping gain, and $x_H$ and $p_H$ are the human LIP's CoM and CoP positions, respectively. $\left( x_H - p_H \right)$ represents the human LIP's tilt. The higher the gain, the faster the robot will travel with a certain human tilt, hence the more sensitive the teleoperation. For implementation, we set $\ddot{x}_{Rcmd} = 0$ and $\zeta_R = 1$. We also compute $x_{Rcmd}$ by integrating $\dot{x}_{Rcmd}$ based on the initial $x_R$ to ensure the coherence between velocity and position commands. 

\subsection{Feedforward Teleoperation Mapping}
FF mapping maps the human LIP's tilt to the robot LIP's tilt. Since an LIP's tilt is proportional to its CoM acceleration, FF mapping can also be understood as mapping the human LIP's tilt to the robot LIP's CoM acceleration. Intuitively, the human body ``becomes" the robot with FF mapping, achieving a sense of dynamic similarity \cite{Ramos_RAL_2018}. On the other hand, the robot does not have any feedback controller, so its stability entirely relies on feedforward control from the human, which gives the mapping's name. 

Mathematically, FF mapping has the expression: 
\begin{align} \label{p_R_FF}
    p_R = x_R - K_{FF} \frac{h_R}{h_H} \left( x_H - p_H \right),
\end{align}
where $h_R$ and $h_H$ are the robot LIP's and human LIP's respective heights, and $K_{FF}$ is the user-selected mapping gain. Rearranging (\ref{p_R_FF}) yields: 
\begin{align}
    \frac{x_R - p_R}{h_R} = K_{FF} \frac{x_H - p_H}{h_H},
\end{align}
which represents the synchronization between the human LIP's and the robot LIP's tilts adjusted by the gain. The higher the gain, the more the robot will tilt, i.e., accelerate, with a certain human tilt, hence the more sensitive the teleoperation. 

Note that FB and FF mappings reflect opposite design philosophies and represent the two extremities in shared control. One allows the robot the stabilize itself and merely regards the human as a command input, whereas the other makes the robot completely dependent on the human. In other words, one grants total control authority to the robot, whereas the other to the human. 

\subsection{Force Feedback Mapping}
To prevent the human from falling during the teleoperation, we designed the HMI feedback force as a spring force proportional to the human LIP's tilt:
\begin{align}
    F_{HMI} = K_{HMI} \left( x_H - p_H \right), 
\end{align}
where $F_{HMI}$ and $K_{HMI}$ stand for the HMI feedback force and the virtual spring stiffness, respectively. $K_{HMI}$ is dependent on the human LIP's height. It takes the numerical value such that LISA will exert its 100 N maximum force when the human tilts to the robot's prescribed maximum tilt angle.

\section{Experimental Design}

\begin{figure*}[t]
\centering
    \includegraphics[width = 0.999\linewidth]{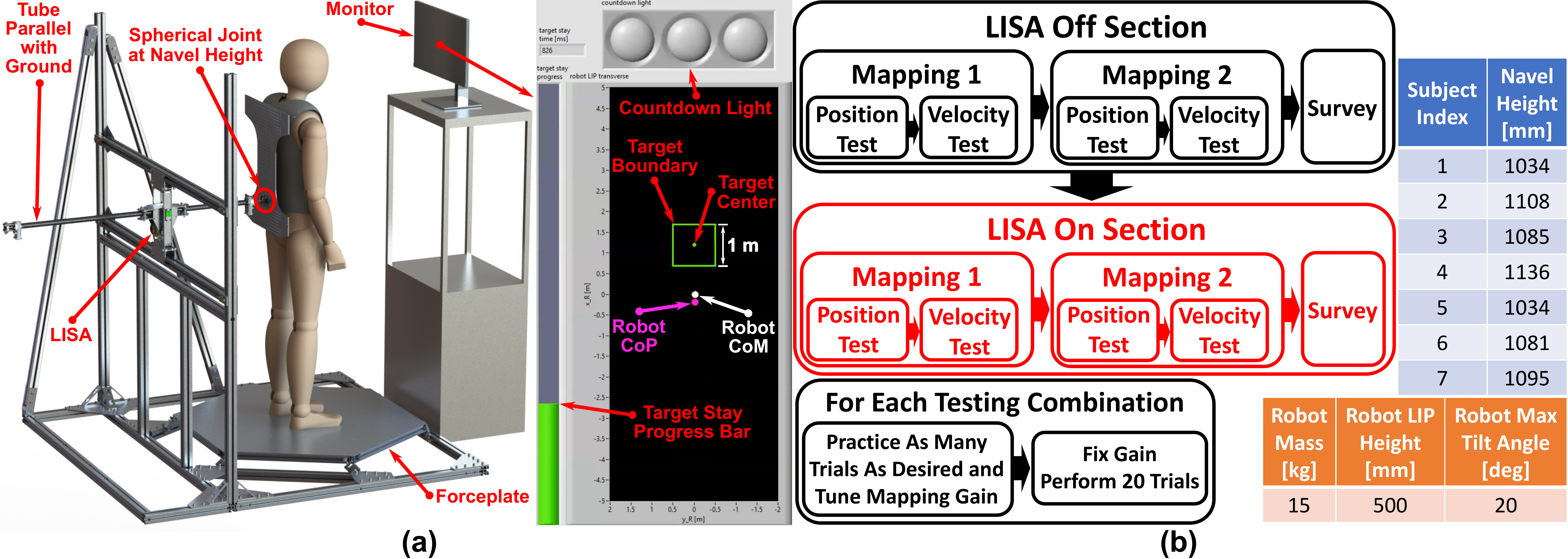}
    \caption{(a) Experimental setup and the graphical interface. (b) Experimental process for each subject and constant parameters during the experiment. }
    \label{fig:experimental_setup_and_process}
\end{figure*}

The experiment's objectives are to compare FB and FF mappings and evaluate the force feedback's effect. Inspired by works in human-computer interaction and neuromechanics \cite{Fitts_1954, Hogan_Dynamic_Primitive}, we developed a setup and a human subject experiment using a simulated LIP as the robot. The robot's mass and height are 15 kg and 0.5 m, respectively. To allow the robot to move dynamically, we set its maximum tilt angle as $20\degree$, i.e., $|\theta_R| \leq 20 \degree$ in Fig. \ref{fig:LIP}. $\theta_H$ does not have any constraint. 

As shown in Fig. \ref{fig:experimental_setup_and_process}.(a), the setup consists of the HMI, a monitor raised to human eye level, and a graphical interface. The interface is in the robot body frame's top view, and shows the robot's CoM, CoP, and a rectangular target. We designed two tests based on this interface: 1) Position test. 2) Velocity test. In position test, the target will stand still at a random distance between 3--5 m in front of the robot. In velocity test, the target will appear at a random distance between 1--2 m in front of the robot, and then move away from the robot at a random constant velocity between 2--4 m/s after a three-second countdown. For both tests, the human's task is to teleoperate the robot after the countdown such that the robot's CoM stays within the target for three seconds. The human must complete the task as fast as possible. An NI cRIO-9082 real-time computer executes the simulation and experiment at 1 kHz, and logs data at 200 Hz. The following video demonstrates the experiment in action: \href{https://youtu.be/GRI0GLWt-hs}{\textcolor{blue}{\underline{https://youtu.be/GRI0GLWt-hs}}}. The University of Illinois at Urbana-Champaign Institutional Review Board has reviewed and approved this research study. 

\begin{figure}[t]
\centering
    \includegraphics[width = 0.999\linewidth]{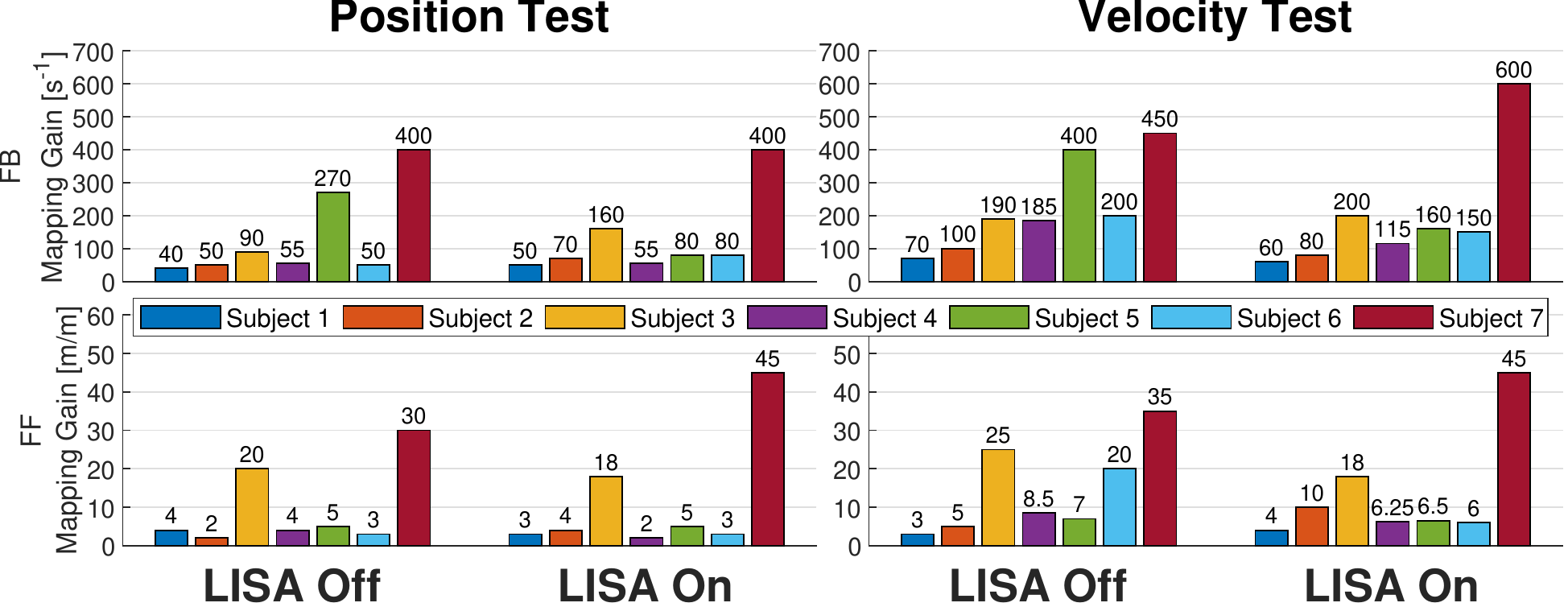}
    \caption{Mapping gains selected by every subject in every combination. }
    \label{fig:mapping_Gain}
\end{figure}

We recruited six male and one female subjects for the experiment. The seven subjects' age mean and standard deviation are 27.86 and 3.08 years, respectively. We used the subjects' navel height as their CoM height. Fig. \ref{fig:experimental_setup_and_process}.(b) shows the experimental process for each subject. In LISA-off and LISA-on sections, LISA's actuation is off and on, respectively. Mapping 1 is FB mapping for subjects 1--3 and FF mapping for subjects 4--7. With two sections, two mappings, and two tests, each subject undergoes eight testing combinations (e.g., LISA-off section + FB mapping + position test is one combination). For each combination, the subject will first practice for however many trials he/she desires and tune the mapping gain. During velocity test practice, the target's velocity is always 4 m/s. After the practice, the gain will be fixed, and the subject will perform 20 trials. At the end of each section, the subject will complete a survey involving: 1) The NASA TLX \cite{NASA_TLX}. 2) Choosing the preferred mapping. 3) Ranking the section's four combinations in difficulty. 4) Commenting subjectively. The subject will rest between the two sections. Excluding the practice, each section consumes about two hours. 

\begin{figure*}[t]
\centering
    \includegraphics[width = 0.999\linewidth]{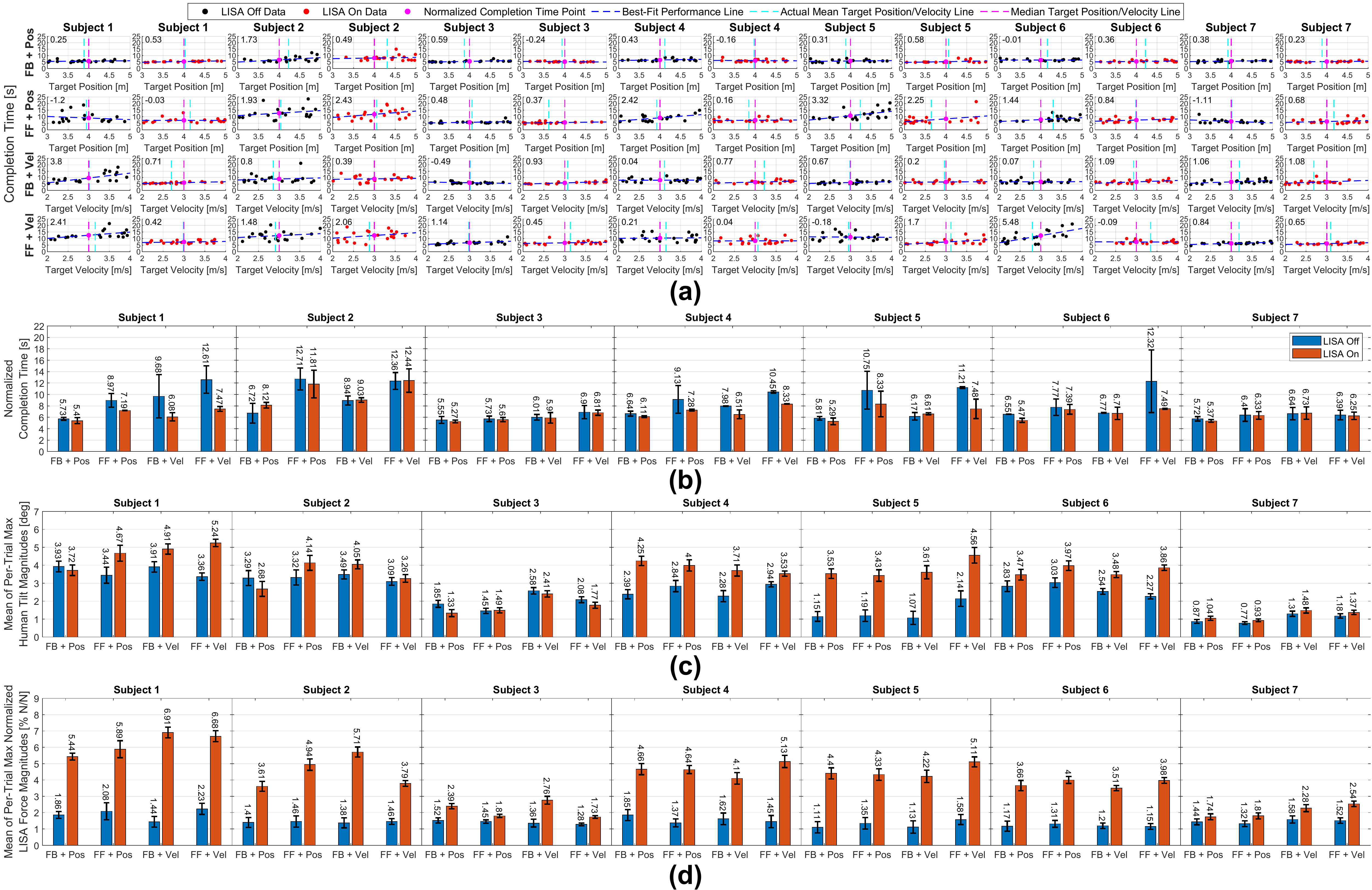}
    \caption{(a) Every subject's completion time VS target position/velocity plots for every combination. The numerical value on each subplot's top left corner is the best-fit performance line's slope. (b) Normalized completion time and deviation. (c) Maximum human tilt. (d) Maximum normalized LISA force.}
    \label{fig:experimental_results}
\end{figure*}

\section{Experimental Results and Discussion}

\subsection{Normalized Completion Time}
The most representative experimental result is the combination completion time, as shown in Fig. \ref{fig:experimental_results}.(a). Since the target position and velocity are random in every trial, and a farther and faster target will lead to a longer completion time, we normalized every subject's per-trial completion times. Specifically, inspired by Fitts's law \cite{Fitts_1954}, we used the linear best-fit line of the 20 data points on the per-trial completion time VS target position/velocity plot to represent a subject's performance in one combination. We then computed the best-fit line's y-coordinate at the x-coordinate of the median of possible target position and velocity, i.e., 4 m and 3 m/s. That y-coordinate is the normalized completion time. The deviation is the absolute difference between the best-fit line's y-coordinates at the two ends of the target position/velocity range. Fig. \ref{fig:experimental_results}.(b) summarizes the normalized results. High performance is characterized by short normalized completion time and small deviation, i.e., small absolute value of the best-fit line's slope. The results suggest that most subjects performed better with FB mapping and when the force feedback was on. 

\subsection{Teleoperation Style Difference}
Though most subjects performed better with the force feedback, some subjects experienced less performance variation after the force feedback became available. Particularly, subjects 3 and 7 exhibit consistently high and stable performances compared with other subjects. In fact, the force feedback appears to have benefited subjects 3 and 7 little, but they are the two best-performing subjects in five out of eight combinations in terms of normalized completion time. A closer look reveals that these two subjects chose higher mapping gains, especially with FF mapping, as shown in Fig. \ref{fig:mapping_Gain}. 

\begin{figure*}[t]
\centering
    \includegraphics[width = 0.999\linewidth]{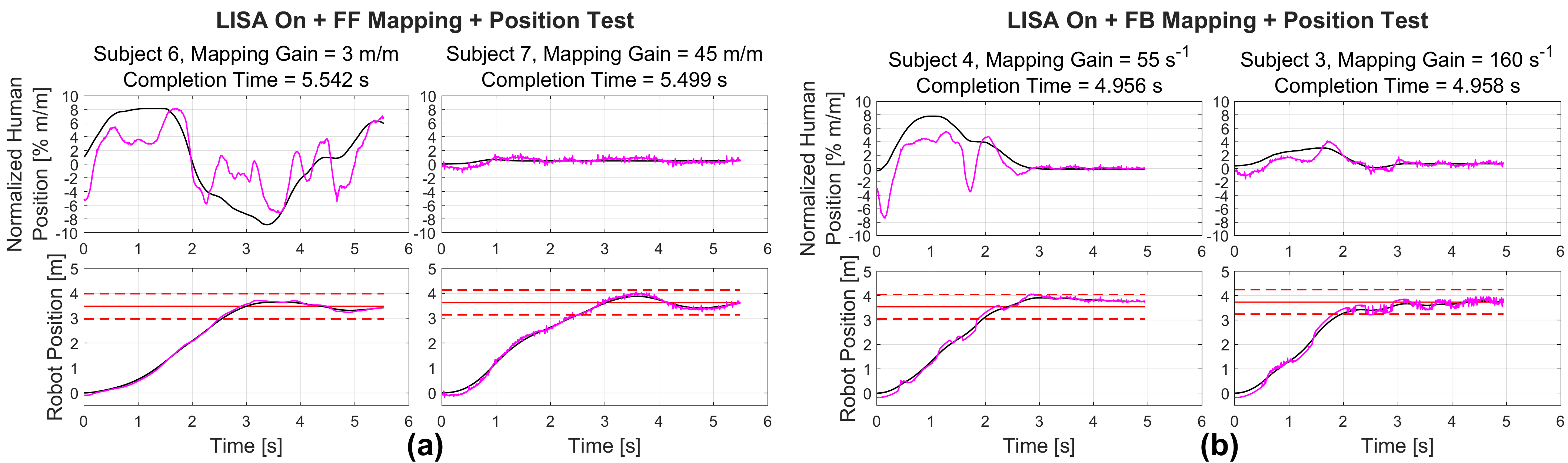}
    \caption{Teleoperation style comparison for FF mapping (a) and FB mapping (b) in one trial. The human postions are normalized by human CoM height. }
    \label{fig:style_comparison}
\end{figure*}

Following this evidence, we computed every subject's mean and standard deviation of per-trial maximum human tilt magnitudes for every combination using the expression: $\text{tilt} = \text{atan} \left( \frac{\left| x_H - p_H \right|}{h_H} \right)$. As shown in Fig. \ref{fig:experimental_results}.(c), the results indicate that subjects 3 and 7 tilted less than other subjects did in most combinations. They also did not tilt more drastically in LISA-on section than in LISA-off section as other subjects did. Hence, we summarize two distinct teleoperation styles: 1) Low gain + large tilt (LGLT) style, represented by all subjects except subjects 3 and 7. 2) High gain + small tilt (HGST) style, represented by subjects 3 and 7.

Fig. \ref{fig:style_comparison} illustrates comparisons between the two styles in one trial. As shown in Fig. \ref{fig:style_comparison}.(a), with FF mapping, subject 7's mapping gain was 15 times of subject 3's, but both subjects achieved similar completion times and robot trajectories. Yet, subject 6's normalized human CoM and CoP trajectories have larger amplitudes and are smoother than subject 7's. Subject 6's robot CoP trajectory is also smoother than subject 7's. This phenomenon is consistent with the mapping design that a higher gain corresponds to more sensitive teleoperation. Since the robot CoP position is the input, which is discontinuous in general, sensitive teleoperation may cause jittery robot CoP trajectory, as shown in subject 7's case. Similar phenomenon also occurred with FB mapping, as in Fig. \ref{fig:style_comparison}.(b). 

We have two assumptions about why such teleoperation style difference exists. The first one is that higher mapping gains reflect higher teleoperation proficiencies. The reasoning is that an HGST-style subject likely possesses motor skills that enable him/her to react fast enough to the sensitive teleoperation, whereas an LGLT-style subject might not. Specifically, high-gain teleoperation requires more jerky movements and greater actuation efforts than low-gain teleoperation. Hence, the LGLT style, which most subjects chose, could be a natural tendency to minimize jerk and effort \cite{Hogan_jerk}. In this regard, the HGST-style subjects overcame their natural tendency and employed more advanced motor skills. Since the robot can attain higher top speed and accelerate faster with higher gains, the HGST-style subjects chose high gains to minimize the time for the robot to catch the target and the completion time instead of jerk or effort. They did so because they could handle the more demanding teleoperation, which an LGLT-style subject might not be able to handle. This assumption is supported by the subjects' normalized completion time ranking. 

The second assumption is that teleoperation style is a subjective preference. The LGLT-style subjects had long normalized completion times because a robot with low gains could not move fast enough to catch the target as quickly as a robot with high gains. This was especially true for combinations with FB mapping and velocity test, where the mapping gain determines the robot's top speed. As shown in Fig. \ref{fig:mapping_Gain}, with FB mapping, all subjects selected higher gains in velocity test than in position test. This assumption is also consistent with control rate setting in model aircraft aerobatics, which depends on the maneuvers the aircraft is performing and the pilot's preference \cite{model_aircraft_dual_rate}. 

\subsection{Robot CoP Trajectory During FB Mapping}
Although the teleoperation style difference is consistent with both teleoperation mappings, we observed two features that are exclusive for FB mapping: 1) The robot CoP trajectory has discontinuous snaps. 2) The human's and the robot's CoP trajectories become more jittery when the robot moves slowly. Both features exist for subjects with both teleoperation styles. 

The root cause of both features is that the robot CoP position is the discontinuous input. Since the robot has its own feedback controller with FB mapping, the human cannot directly control the robot CoP position. Instead, the robot controls it to track the human's velocity command, which may not embed dynamic similarity with the human or follow the relatively smooth human CoP trajectory. 

In addition, feature 1) involves the maximum tilt angle imposed on the robot. By (\ref{LIP_dynamics}), this constraint is equivalent to limiting the maximum robot CoM acceleration. Hence, there could be moments when the robot cannot catch up with the change of the human's velocity command even at its maximum acceleration. In such cases, the robot CoP will abruptly move to and stay at its maximum relative to the CoM, which causes the snaps in the robot CoP trajectory in Fig. \ref{fig:style_comparison}.(b). 

Feature 2) is a sign that the robot is rapidly switching between acceleration and deceleration. The robot does so because the human rarely commands a perfectly constant velocity, but subtly and continuously adjusts his/her tilt to recover from overshoot and accelerate/decelerate the robot. Because the robot CoP must move beyond the CoM to produce deceleration, the robot CoP trajectory appears as a high-frequency oscillation around the robot CoM trajectory. The oscillation's frequency is positively correlated to the mapping gain, which is demonstrated by the comparison between the two subjects' robot trajectories in Fig. \ref{fig:style_comparison}.(b). 

\subsection{Normalized HMI Feedback Force}
Fig. \ref{fig:experimental_results}.(d) shows every subject's mean of per-trial maximum normalized LISA force magnitudes. We normalized LISA force by dividing it by the subject's body weight. Consistent with the maximum tilt results in Fig. \ref{fig:experimental_results}.(c), the HGST-style subjects have smaller normalized LSIA forces than the LGLT-style subjects, since LISA force was a spring force proportional to the human's tilt. Moreover, for all subjects' LISA-off sections, the maximum normalized LISA force is between about $1\%$--$2.3\%$, which reflects LISA's high backdrivability. 

\begin{figure}[t]
\centering
    \includegraphics[width = 0.999\linewidth]{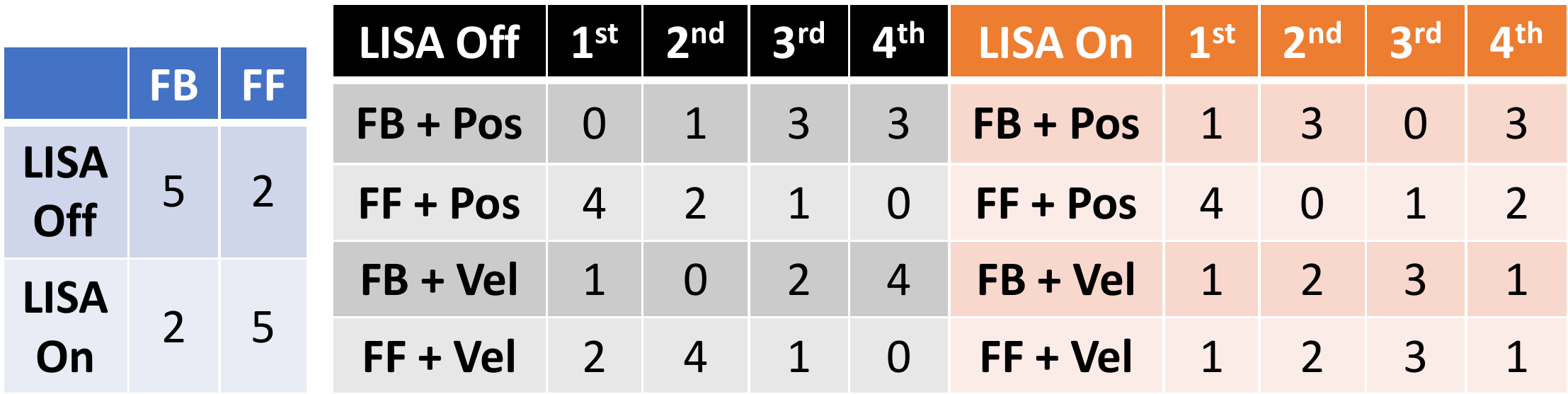}
    \caption{Survey results of preferred mapping (left) and combination difficulty ranking (right). $1^{\text{st}}$ is the most difficult and $4^{\text{th}}$ the easiest. Numerical value in each entry represents number of subjects. }
    \label{fig:survey}
\end{figure}

\subsection{Survey Results}
For the NASA TLX, most subjects reported lower demands, effort, and frustration, and higher performance for LISA-on section than for LISA-off section. All subjects considered the force feedback to be helpful for the teleoperation. Specifically, all subjects commented that the force feedback made them feel more secure when they tilted their body. Six subjects commented that the force feedback enabled them to find their zero tilt angle better and move their CoP more quickly and drastically, especially when they tilted backward. Four subjects mentioned that the force feedback reduced the physical strains the tilting caused on their toes and heels. Two subjects mentioned that, with FB mapping, the robot sometimes ``fought against" the human command and was not as responsive as with FF mapping. They also suggested that FB mapping might be more suitable for smooth and gradual maneuvers, whereas FF mapping more suitable for swift and dynamic maneuvers. Furthermore, the two HGST-style subjects acknowledged that they did not benefit significantly from the force feedback. In fact, one of them teleoperated the robot by pressing different parts of the feet to the forceplate without moving the upper body in some trials.  

Fig. \ref{fig:survey} shows the subjects' mapping preferences and combination difficulty rankings, from which we observed the following: 1) The force feedback appears to have changed the subjects' mapping preferences, as more subjects preferred FF mapping when the force feedback was on. They did so even though most of them achieved shorter normalized completion times with FB mapping in LISA-on section. 2) More subjects considered velocity test with FF mapping to be easier when the force feedback was on. However, FF mapping + position test remains the most difficult for most subjects. 

\subsection{Limitations of This Study}
This study's primary limitation is the relatively small number of subjects due to the experiment's time consumption and strenuousness. Moreover, the target velocities during the experiment are relatively high considering the simulated robot's size. This part of experimental design might have been biased toward the HGST-style subjects, since the LGLT-style subjects could not teleoperate the robot to catch the target as fast as the HGST-style subjects could. Finally, the experiment was based on a simulated robot, which assumes a linear model and perfect sensing and actuation. These idealizations will break down during hardware implementation, so the teleoperation's practicality must be verified on a real robot. 

\section{Conclusions and Future Work}
The contributions of this work are: 1) To introduce the HMI's design and performance. 2) To demonstrate and evaluate our concept of wheeled humanoid robot locomotion teleoperation by body tilt using the HMI and a simulated robot via a dynamic target tracking experiment. For contribution 1), we presented LISA's and the forceplate's designs and the analysis on LISA's force-current profile, step response, and reflected inertia. For contribution 2), we examined the subjects' normalized completion times, maximum tilts, maximum normalized LISA forces, and survey responses. The results suggest that most subjects performed better with FB mapping and the force feedback. However, we discovered two teleoperation styles, and the force feedback benefited the HGST-style subjects less than the LGLT-style subjects. We compared the two styles with human and robot CoM and CoP position data from one trial for both mappings, and proposed two assumptions about why the teleoperation style difference exists. We also discussed the oscillations in robot CoP trajectory that occurred exclusively for FB mapping. Lastly, the survey results show that more subjects preferred FF mapping and considered velocity test with FF mapping to be easier when the force feedback was on. Yet, most subjects achieved shorter normalized completion times with FB mapping in LISA-on section, and ranked FF mapping + position test as the most difficult regardless of the force feedback's availability. 

Future works after this study include: 1) To implement the two teleoperation mappings on a real wheeled humanoid robot and evaluate the mappings' performances. 2) To combine the two teleoperation mappings---potentially in state-dependent manners---to preserve the strengths of both mappings while compensating for the weaknesses of each to achieve stable, intuitive, and dynamic locomotion teleoperation. 

\section*{Acknowledgement}
The corresponding author would like to sincerely thank Dillan Kenney and Dr. Yeongtae Jung for their assistance in system integration, Guillermo Colin and Yu Zhou for their dedication in data collection, and all subjects for their commitment and perseverance during the experiment. 

\bibliographystyle{IEEEtran}
\bibliography{ICRA_2022_References.bib}

\end{document}